\documentclass[runningheads]{llncs}
\usepackage{graphicx}
\usepackage{comment}
\begin{document}
\title{Shape Reconstruction from Thoracoscopic Images using Self-supervised Virtual Learning}

\titlerunning{Shape Reconstruction using Self-supervised Virtual Learning}
%
\author{Tomoki Oya\inst{1}
\and
Megumi Nakao\inst{2}
\and
Tetsuya Matsuda\inst{1}
}
\authorrunning{T. Oya et al.}
%
\institute{Graduate School of Informatics, Kyoto University, Kyoto, Japan.
\email{t-oya@sys.i.kyoto-u.ac.jp}
\\
\and Graduate School of Madicine, Kyoto University, Kyoto, Japan.
}

\maketitle

\begin{abstract}
Intraoperative shape reconstruction of organs from endoscopic camera images is a complex yet indispensable technique for image-guided surgery. To address the uncertainty in reconstructing entire shapes from single-viewpoint occluded images, we propose a framework for generative virtual learning of shape reconstruction using image translation with common latent variables between simulated and real images. 
As it is difficult to prepare sufficient amount of data to learn the relationship between endoscopic images and organ shapes, self-supervised virtual learning is performed using simulated images generated from statistical shape models. However, small differences between virtual and real images can degrade the estimation performance even if the simulated images are regarded as equivalent by humans.
To address this issue, a Variational Autoencoder is used to convert real and simulated images into identical synthetic images. In this study, we targeted the shape reconstruction of collapsed lungs from thoracoscopic images and confirmed that virtual learning could improve the similarity between real and simulated images. Furthermore, shape reconstruction error could be improved by 16.9$\%$.

\end{abstract}
\section{Introduction}

Endoscopic camera images have been used to reach organs during surgery. However, it is difficult to grasp the three-dimensional structure of organs inside the body owing to the high uncertainty caused by organ deformation and narrow visibility during surgery. Therefore, research is underway to reconstruct the 3D shape of organs from endoscopic images that are acquired during surgery. In this paper, we focus on the reconstruction of 3D shapes using single-viewpoint endoscopic camera images, which is a key technique for image-guided surgery.

Shape reconstruction of biological organs using endoscopic camera images reported so far are: A study on the recovery of three-dimensional information of the entire stomach including texture from endoscopic images of the stomach\cite{endoscopic_reconstruction}, and model-based positioning of the liver\cite{koo}. The former reconstructs shape information from video, but only recovers the shape of the organ surface within the visible range, and the latter calculates the deformation using parameter optimization, but the computational cost and stability of the optimization calculation remains an issue.
To address these issues, there have been reports on attempts to predict 3D shapes and deformations using statistical shape models and graph convolution networks, which describe the morphological characteristics of biological organs in terms of low-dimensional parameters\cite{sdm}\cite{X2S}\cite{Richard}\cite{multi}.
In the field of medical imaging, data augmentation with simulated data generated from statistical shape models has been shown to be effective.
Tang et al. achieved an improvement in segmentation accuracy with a data augmentation method using statistical shape models compared with a method using real images\cite{ssm}. 
However, it has been reported that if excessive simulated images are used for training, the accuracy decreases in contrast to the increase in training data; hence, the problem of treating simulated and real images equally still remains.

In this paper, we propose a framework for self-supervised virtual learning for the shape reconstruction of deformed organs using image translation with common latent variables between simulated and real images from endoscopic camera images. This method solves the problem of insufficient training data in medical imaging by using statistically generated simulated images. In this paper, we refer to this learning method as ``virtual learning.'' We use a Variational Autoencoder (VAE)\cite{vae} to extract latent variables that are common to both simulated and real images, and construct a framework that accepts two images with different features that humans may regard as equivalent as inputs and converts them into identical synthetic images. 
Although some studies\cite{funsyn} have reported the generation of medical images that represent pathological, morphological, and anatomical variations of VAEs, there have been no reports of their application to simulation-based learning, which is the subject of this study.
Since our framework can improve the similarity between simulated and real images generated from 3D CT images of individual patients, it can reconstruct the shape of lungs from endoscopic images in thoracoscopic lung cancer resection with higher accuracy than conventional methods. 

\section{Method}
\begin{figure}[tb]
      \centering
     	    \includegraphics[scale=0.25]{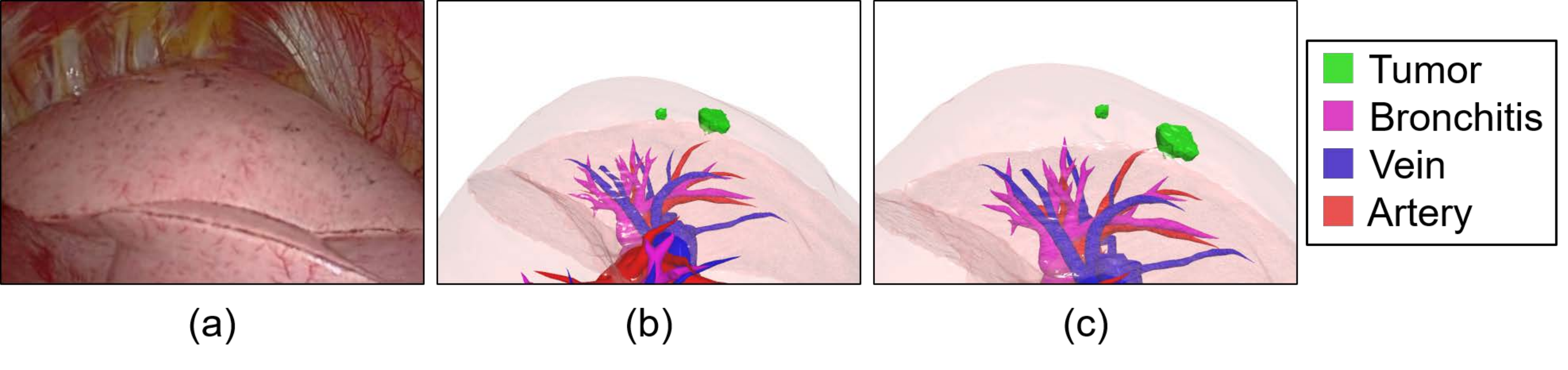}
        \caption{Examples of real and simulated images in endoscopic images, (a) real image $I_R$ extracted from a surgical video, (b) different appearances generated by changing weights, and (c) simulated image $I_S$ corresponding to the real image}
        \label{data_example}
\end{figure}

\subsection{Dataset and Problem Definition}
In this study, 128 still images were collected from videos of of thoracoscopic resections of 9 cases of the right lung and 3 cases of the left lung.
The lung shape $M$ during insufflation was constructed from the CT data taken before the surgery, and the lung shape $\hat{M}$ during collapse was constructed from the cone-beam CT data taken during the surgery. The lung shape is a mesh model consisting of 402 vertices and 796 triangles.

Fig. \ref{data_example} shows an example of an endoscopic image $I_R$ and a simulated image $I_S$ generated from the statistical shape models. In Fig. \ref{data_example}(a), part of the surface of the upper lobe of the lung and the inner wall of the thoracic cavity can be seen. Fig. \ref{data_example}(b) shows an example of reproducing different lung shapes by changing the weights of the principal components of the statistical shape models as well as the camera parameters. The lung statistical shape model is a statistical model with low-dimensional parameters describing the displacement vectors per vertex obtained from the vertex-correlated lung shape meshes obtained by mesh deformation alignment\cite{lung1}\cite{lung2} between $M$ and $\hat{M}$.
Owing to the collapse deformation, the lung shape at the time of air inclusion, which can be obtained before surgery, differs significantly from the lung shape at the time of collapse during surgery. Therefore, the simulated image reproduces lungs with various appearances and solves the problem pertaining to a lack of data, which is a persistent issue in the field of medical imaging. Fig. \ref{data_example}(c) shows the results of manually adjusting each parameter to correspond to the surgical scene under the supervision of a surgeon.
Since the lung surface is transparent and the tumor location, bronchus, and vascular structures can be comprehended, it can be used for surgical assistance. If the parameters of statistical shape models can be automatically calculated using the endoscopic images and the two can be aligned, the shape of organs and anatomical structures can be identified during surgery.

In endoscopic surgery, several situations are considered for the endoscopic images and target organs to be captured during an operation. In this study, we considered the following situations that are commonly observed in most surgeries.
\begin{itemize}
    \item The initial shape $M$ of the organ to be operated on is obtained from the 3D-CT measured before the operation.
    \item A part of the surface of the organ can be observed in the endoscopic image $I_R$, but more than 50 $\%$ of the entire organ is occluded.
    \item Information on the position of the endoscope camera and the direction of the eye are not available.
    \item The organ is deformed during observation, but its deformation is statistically predictable.
\end{itemize}
In this study, we aimed to obtain the 3D shape of an organ under surgery directly from the endoscope image via virtual learning using a simulated image $I_S$ that can be generated from the statistical shape models under the above conditions, and using image translation with common latent variables between the endoscope image $I_R$ and the simulated image $I_S$. The images used in this study were standardized to 180×120 pixels, and 10,000 simulated images generated for each case were used for training.

\subsection{Framework}
\begin{figure}[tb]
  \centering
    	   \includegraphics[scale=0.35]{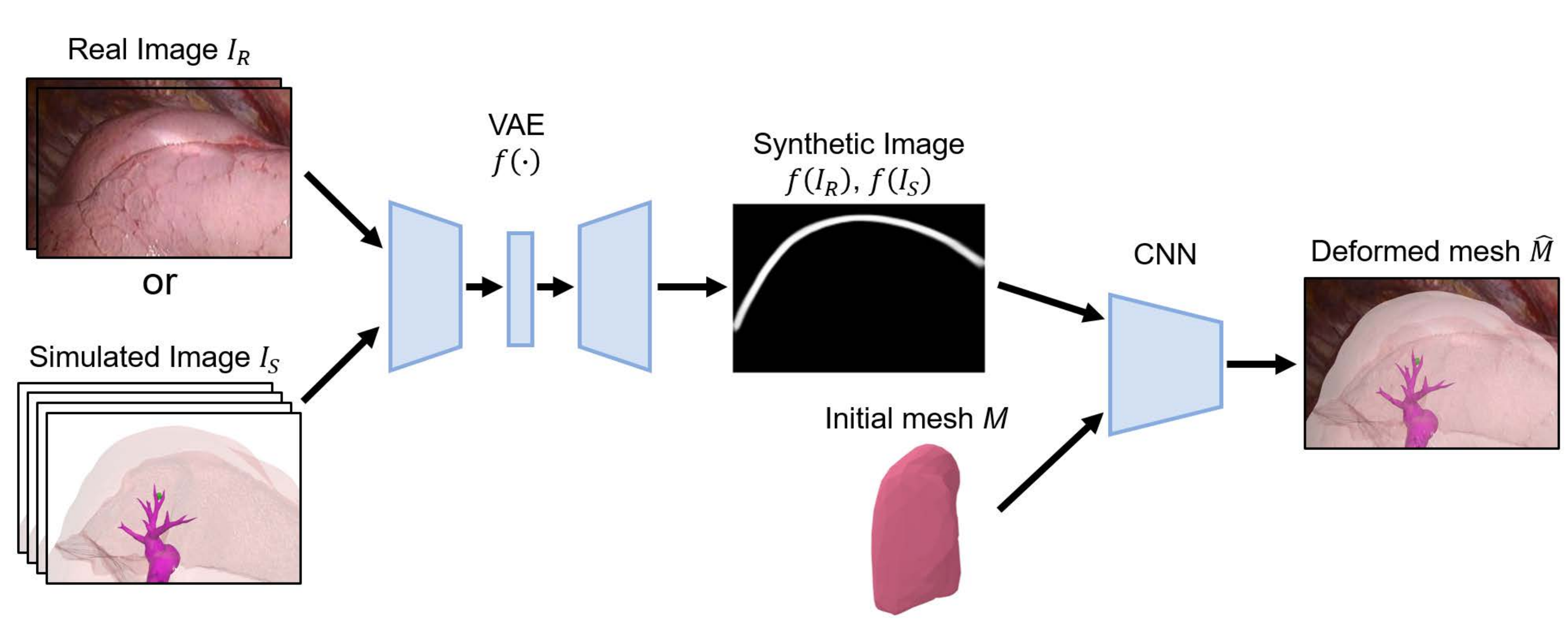}
      		\caption{Proposed Framework. Variational Autoencoder (VAE) trains the common latent variables in the real image $I_R$ and the simulated image $I_S$, while the convolutional neural network (CNN) trains the relationship between the synthetic image and the model parameters $\theta$.}
      \label{framework}
\end{figure}

Fig. \ref{framework} shows the proposed framework, which consists of a VAE to achieve image transformation with common latent variables between real and simulated images, a convolution neural network (CNN) to learn the relationships between synthetic images and model parameters. 
In the VAE, the real $I_R$ and the simulated $I_S$ images, which have different features but are regarded as equivalent by humans, are transformed into an identical synthetic image.  
Then, in the CNN, we perform self-supervised virtual learning, using the synthetic image and the initial shape $M$ of the organ as input, and outputting the deformation mesh and camera parameters.
The details of the VAE are described in Section \ref{seq:vae}, and the details of the CNN are described in Section \ref{seq:vl}.

\subsection{Image translation}
\label{seq:vae}
We propose a transformation function $f$ that translates real and simulated images, which have different detailed features but are regarded as equivalent by humans, into an identical synthetic image. In this study, the input images to VAE are the real image $I_R$ and the simulated image $I_S$, which is an artificial representation of $I_R$. The transformation function $f$ plays the role of translating $I_R$ to $I_S$ and reconstructing $I_S$ to $I_S$.

VAE consists of an encoder, which converts input data $x$ into latent variables $z$, and a decoder, which reconstructs the same input data $x$ from the latent variables $z$. The encoder and decoder in our study consists of three fully connected layers.  Let $\phi$ and $\theta$ be the parameters in the neural networks of the encoder and decoder, and let $q_{\phi}(z|x)$ and $p_{\theta}(x|z)$ be the probability distributions of the encoder and decoder, respectively. $p(x)$ is a Gaussian distribution and the encoder learns the parameter $\phi$ for $q_{\phi}(z|x)$, which generates parameters $\mu$ and $\sigma$ of $p(x)$, and the decoder learns the parameter $\theta$ for $p_{\theta}(x|z)$, which reconstructs $x$ from $z$. 

Since $I_R$ and $I_S$ are compressed into a common latent variable, and $f(I_R)$ and $f(I_S)$ are output from the common latent variable, we assumed that it would be possible to reconstruct the identical synthetic image from the common latent variable; therefore, we adopted VAE in this study. Let $g$ be the transformation function of VAE in the usual learning method\cite{vae2}. When the real image $I_R$ is input into the variable function $g$, the relationship between the input $I_R$ and the output $g(I_R)$ is $g(I_R) \approx I_R$. When the simulated image $I_S$ is input into the variable function $g$, the relationship between the input $I_S$ and the output $g(I_S)$ is $g(I_S) \approx I_S$. By contrast, in this study, when the real image $I_R$ is input into the transformation function $f$ of VAE, the relationship between the input $I_R$ and the output $f(I_R)$ is $f(I_R) \approx I_S$. When the simulated image $I_S$ is used as input, the relationship between the input $I_S$ and the output $f(I_S)$ is $f(I_S) \approx I_S$. By using the transformation function $f$ to transform two images with different features into the identical synthetic image, we assumed it would be possible to learn that $I_R$ and $I_S$ are equivalent. In addition, synthetic images $f(I_R)$ and $f(I_S)$, which are transformed from $I_R$ and $I_S$, are highly similar to each other.

In VAE, the error between the input image to the encoder and the image reconstructed by the decoder must be minimized. In addition, a regularization term that minimizes the error between the distribution of $q_{\phi}(z|x)$ estimated by the encoder and $p_{\theta}(z)$, which is the prior distribution of $z$(both Gaussian), is required. Therefore, in VAE training, we introduce a loss function that is a weighted sum of the reconstruction error shown in the first term of Eq. (\ref{eq:vae_loss}) and the Kullback-Leibler Divergence (KLD) shown in the second term of Eq. (\ref{eq:vae_loss}).

\begin{equation}
    \mathcal{L} = \|x-\hat{x}\|^2_2+ \lambda_{KL} D_{KL}(q(z|x)\parallel p(z))
    \label{eq:vae_loss}
\end{equation}

\subsection{Virtual learning}
\label{seq:vl}
In this section, we describe virtual learning using the simulated images that can be generated statistically. In statistical shape models, the collapse deformations assumed during surgery can be generated from the preoperative CT models of individual patients before surgery by changing the weights related to the lung collapse volume. The organ deformations are based on the average collapsed deformations by learning. 
The vertex coordinates $\upsilon_i^{'}$ of the generated $\hat{M}$ can be obtained by the Eq. (\ref{eq:deform}).

\begin{equation}
    \upsilon_{i}^{'} = \upsilon_i + \bar{u} + \omega \cdot u_i 
    \label{eq:deform}
\end{equation}
where $\upsilon_i$ is the position of the vertex of the lung shape during aeration, $\bar{u}$ is the average displacement due to collapse deformation, and $u_i$ is the displacement vector obtained from the lung statistical shape model. $\omega$ is the weight of the displacement, where $\omega < 1.0$ represents a squeezed shape compared with the average displacement due to collapse deformation, and $\omega > 1.0$ represents a bulged shape compared with the average displacement due to collapse deformation.
In this study, the parameters were varied with a width of 0.1 around the average displacement due to collapse deformation ($\omega = 1.0$).

Next, $\hat{M}$ is rendered with the set camera parameters to generate the simulated image $I_S$. The camera parameters that are expected to change during the surgery are the camera position in three dimensions and the focus point in two dimensions, which is a total of five dimensions.
In this study, a large number of simulated images IS were generated by changing the parameters with a width of 15 mm in the depth direction of the camera position, 25 mm in the other two directions, 15 mm in the horizontal and vertical directions of the focus point for the right lung case, and 10 mm in the three directions of the camera position and two directions of the focus point mentioned above for the left lung case, centering on the camera parameters assumed in advance by the surgeon.
To implement the image transformation described in Section \ref{seq:vae} accurately, in this study, $I_S$ was converted into a binary image of the lung region and the background, and then into a lung contour image by morphological transformation\cite{morph}.

\subsection*{Reconstruction Error}
Since the purpose of this study was to obtain the 3D shape of an organ adjusted to an endoscope image, it was desirable to minimize not only the difference between the estimated shape and the target shape but also the deviation of the position and orientation of the obtained shape. In the proposed method, Reconstruction Error, defined in Eq. (\ref{eq:loss1}), is introduced as a loss function to calculate the error between mesh vertices after projection.

\begin{equation}
    \mathcal{L}_{R} = \frac{1}{n}\sum_{i=1}^{n}\| M_{\theta}\upsilon_{i} - M_{\hat{\theta}}\hat{\upsilon_{i}}\|
    \label{eq:loss1}
\end{equation}
where $M_{\theta}$ and $M_{\hat{\theta}}$ are the target and estimated projection matrices, respectively, $\upsilon_{i}$ and $\hat{\upsilon_{i}}$ are the target and estimated mesh vertex points, respectively, and $M$ is a $4\times4$ matrix used for the perspective projection transformation in generating the rendering image of the organ mesh with camera parameters.

\subsection*{Parameter Loss}
Since the space in the thoracic cavity is limited, the motion of the endoscopic camera is restricted within a certain range. To constrain the range of variations in the camera parameters, the proposed method introduces the Parameter Loss defined in Eq. (\ref{eq:loss2}) as a loss function to calculate the error between the estimated value $\hat{\theta}$ and the true value $\theta$ of the camera parameters and the displacement $\omega$ in the collapse deformation of the lung.

\begin{equation}
    \mathcal{L}_{P} = \frac{1}{n}\sum_{i=1}^{n}| \theta_{i} - \hat{\theta}_{i}|^2_{2}
    \label{eq:loss2}
\end{equation}

The loss function $\mathcal{L}$ for the entire virtual learning is defined as a weighted linear sum of two loss functions as Eq. (\ref{eq:vl}):

\begin{equation}
    \mathcal{L} = \mathcal{L}_{R} + \lambda_{P}\mathcal{L}_{P}
    \label{eq:vl}
\end{equation}

\section{Experiments}
By comparing shape reconstruction methods, the effectiveness of the proposed framework in organ shape reconstruction using a single endoscope image was confirmed. The entire network was implemented in Python 3.6.8, Keras, and TensorFlow GPU libraries.
Virtual learning was performed using the Adam optimizer with the following parameters: batch size = 60, number of epochs = 500, dropout rate = 0.5, and learning rate = $1.0 \times 10^{-3}$ . A single NVIDIA GeForce RTX 2070 GPU was used for training, and the virtual learning took 1.4 hours.

The implementation of this framework enables us to output 3D shapes aligned to the endoscope image from the initial shape. In this study, mean absolute error (MAE), defined as the shape reconstruction error shown in Section \ref{seq:vl}, was used as an evaluation metric for shape reconstruction accuracy. MAE evaluates the average error between 3D shape vertices in the camera coordinate system.
Based on the comparison conditions used in the study by Tang et al., this study compared the results using only real images \cite{ssm}, using simulated images generated from statistical shape models (virtual learning) \cite{ssm}, and synthetic images obtained by transforming simulated images using VAE(the proposed method). As a combination of the loss functions, $\lambda_{KL}=5$, $\lambda_{P}=0.5$ were adopted based on the results of several sets of experiments.

\begin{table}[b]
  \caption{Shape reconstruction error for the proposed method (Proposed), real image only (Real), and virtual learning (Virtual). The
mean $\pm$ standard deviation of MAE [$\mathrm{mm}$].}
  \label{tb:error}
  \begin{center}
    \begin{tabular}{cccccccc}
      \hline
        &  Real  &    &  Virtual  &    &  Proposed  &  \\
      \hline
        & 75.7 $\pm$ 33.6 &  & 27.2 $\pm$ 15.5 &  & 22.6 $\pm$ 12.7&\\
    \hline
    \end{tabular}
  \end{center}
\end{table}

Cross-validation was performed for all 12 cases except for one test case. Table \ref{tb:error} shows the evaluation results of the shape reconstruction in 11 cases, excluding Case 12 owing to the small amount of data. As a result, it can be confirmed that the shape reconstruction error of this method is significantly smaller than that of the method based on Tang et al.'s idea (one-way analysis of variance, ANOVA;
$p < 0.05$ significance level).

\begin{figure}[tb]
  \centering
 	   \includegraphics[scale=0.35]{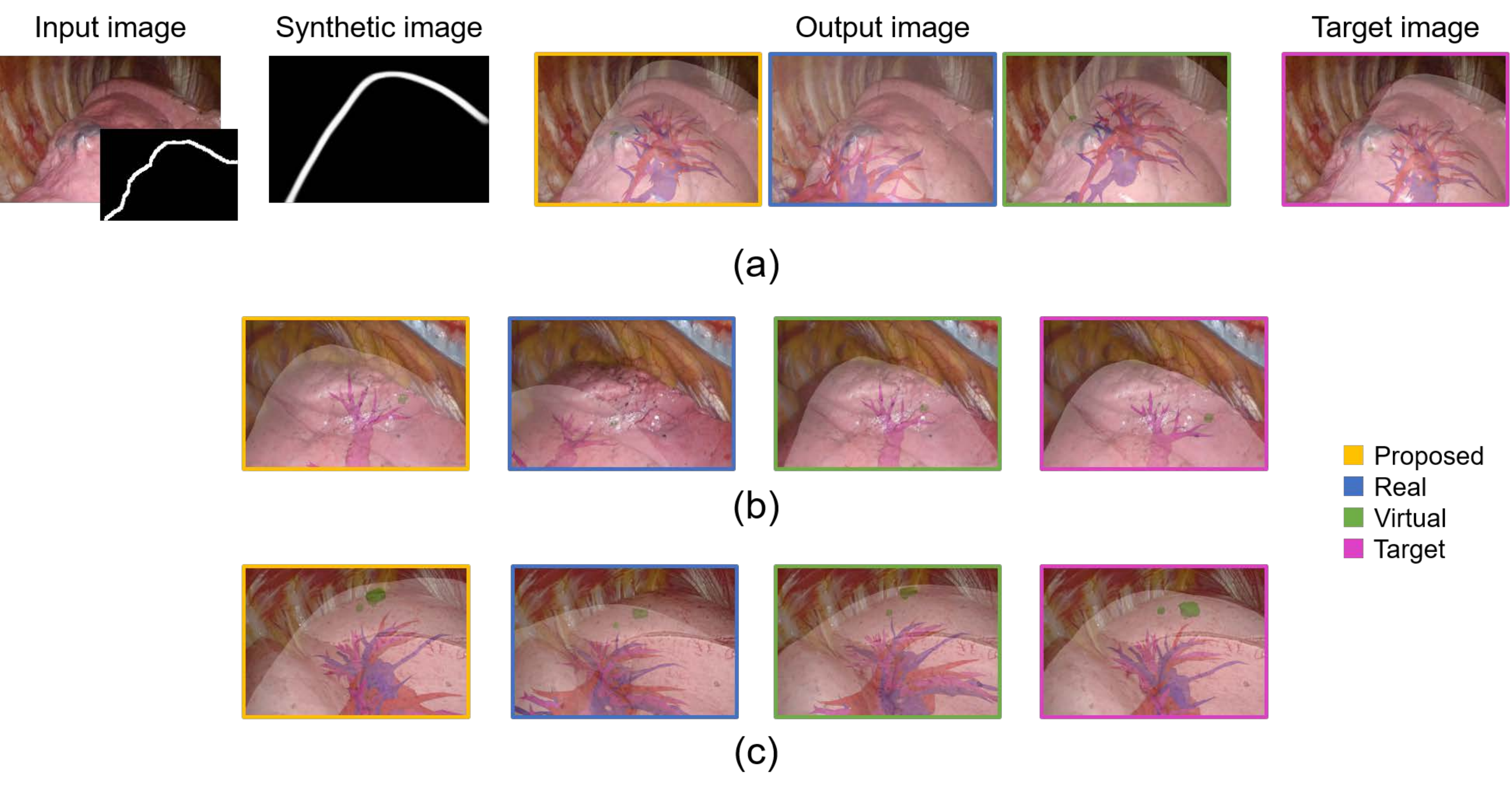}
 \caption{Shape reconstruction results for the proposed method (Proposed), real image only (Real), and virtual learning (Virtual). The proposed method can more accurately estimate the tumor position, vascular structure, and posture.}
 \label{render}
\end{figure}

Fig. \ref{render} shows the visualization of the shape reconstruction results. In Fig. \ref{render}(a), the tumor location and vascular structure of the 3D image of the proposed method are closer to the target image than those of other methods. The MAE of the method using real images only is 112.6 mm, virtual learning is 18.7 mm, and the proposed method is 9.7 mm.
Fig. \ref{render}(b) shows that the output image of the proposed method is closer in appearance to the target image. Fig. \ref{render}(c) shows that the pose of the 3D image is closer to the target image than the other methods; however, there is a shift in the tumor position. Virtual learning can reduce the vertex-level error. In addition, the proposed image transformation method enables the constraint of camera parameters to be applied to the real image, and thus an appropriate pose can be estimated.

\section{Conclusion}
In this paper, we proposed a framework of self-supervised virtual learning for the problem of organ shape reconstruction from endoscopic camera images. Virtual learning based on simulated images generated from statistical shape models was used to cope with the small amount of training data, and the transformation to a common generated image with common latent variables between real and simulated images was realized using VAE. However, the variations in real images are limited, and image transformation results are varied. In the future, we plan to improve the performance of image transformation.

%
%

\newpage

%
%

\end{document}